# Marine chlorophyll prediction and driver analysis based on LSTM-RF hybrid models

Zhouyao Qian[#]
School of Computer and Artificial Intelligence
Nanjing University of Science and Technology ZiJin College
Nanjing,China

Yang Chen[#]
School of Computer and Artificial Intelligence
Nanjing University of Science and Technology ZiJin College
Nanjing,China

Baodian Li
School of business
Nanjing University of Science and Technology ZiJin College
Nanjing,China

Shuyi Zhang
Department Of Electronic engineering & Optoelectronic tech
Nanjing University of Science and Technology ZiJin College
Nanjing,China

Zhen Tian
James Watt School of Engineering
University of Glasgow
Glasgow,United Kingdom

Gongsen Wang
State Key Laboratory of Transducer Technology, Aerospace Information Research Institute
University of Chinese Academy of Sciences
Beijing,China

Tianyue Gu
Department Of Electronic engineering & Optoelectronic tech
Nanjing University of Science and Technology ZiJin College
Nanjing,China

Xinyu Zhou
Department Of Electronic engineering & Optoelectronic tech
Nanjing University of Science and Technology ZiJin College
Nanjing,China

Huilin Chen
Department Of Electronic engineering & Optoelectronic tech
Nanjing University of Science and Technology ZiJin College
Nanjing,China

Xinyi Li
School of business
Nanjing University of Science and Technology ZiJin College
Nanjing,China

Hao Zhu
School of Cyber Science and Engineering
Nanjing University of Science and Technology
Nanjing,China

Shuyao Zhang
School of Humanities and Social Sciences
Nanjing University of Science and Technology ZiJin College
Nanjing, China

Zongheng Li
Department Of Electronic engineering & Optoelectronic tech
Nanjing University of Science and Technology ZiJin College
Nanjing,China

Siyuan Wang[*]
Department Of Electronic engineering & Optoelectronic tech
Nanjing University of Science and Technology ZiJin College
Nanjing,China
19852225218@163.com

[#]These authors contributed equally.

*Abstract*—Marine chlorophyll concentration is an important indicator of ecosystem health and carbon cycle strength, and its accurate prediction is crucial for red tide warning and ecological response. In this paper, we propose a LSTM-RF hybrid model that combines the advantages of LSTM and RF, which solves the deficiencies of a single model in time-series modelling and nonlinear feature portrayal. Trained with multi-source ocean data (temperature, salinity, dissolved oxygen, etc.), the experimental results show that the LSTM-RF model has an $R^2$ of 0.5386, an MSE of 0.005806, and an MAE of 0.057147 on the test set, which is significantly better than using LSTM ($R^2 = 0.0208$) and RF ($R^2 = 0.4934$) alone , respectively. The standardised treatment and sliding window approach improved the prediction accuracy of the model and provided an innovative solution for high-frequency prediction of marine ecological variables.

# I. INTRODUCTION

Marine chlorophyll content is an important indicator for assessing marine ecological health and carbon cycle functioning, which is crucial for marine ecological monitoring, environmental protection and resource management [1-3]. Existing monitoring techniques, such as satellite remote sensing [4] and ship surveys [5], suffer from insufficient spatial and temporal coverage, data latency, and high cost, which limit real-time monitoring and response of marine ecosystems. With the rapid development of machine learning and deep neural network technology, marine ecological modelling has ushered in new opportunities [6-7]. However, single models, such as Long Short-Term Memory (LSTM) [8] and Random Forest (RF) [9] models, have limitations in time-series dependency modelling and nonlinear relationship modelling. How to fuse multi-source data and construct integrated models to improve prediction accuracy has become the key issue to marine chlorophyll dynamics prediction.

In 2012, Doney [10] revealed that the chlorophyll concentration in the North Atlantic Ocean was negatively correlated with water temperature (SST) and positively correlated with nitrate through a generalised additive model (GAM), but did not consider the dynamic process; in 2015, Zhongping Lee [11]proposed an improved OCx algorithm to enhance the accuracy of chlorophyll inversion in turbid waters, but it was difficult to deal with spatio-temporal variations and mutation factors; in the same year, Marie-Hélène Rio [12] combined multi-satellite data and statistical models to construct a global chlorophyll monitoring framework, but it lacked the support of socio-economic variables and had limited ability to generalise across regions; in 2021, Hang Xin [13]screened the key factors of chlorophyll in Lake Taihu by using RF and verified the importance of variable interactions, but relied on a single algorithm, which was difficult to meet the conditions of high-dimensional dynamic ecological modelling needs.

Based on the previous research, the "LSTM-RF hybrid modelling approach" proposed in this study uses LSTM to capture the time-series dynamic features, combines with RF for high-dimensional feature screening and nonlinear pattern modelling, and at the same time introduces the attention mechanism to optimize the feature weight allocation, so as to enhance the model's sensitivity to the key driving factors, and the LSTM-RF model improves the prediction through the dynamic integration of multiple sources. The LSTM-RF model improves the comprehensiveness of the prediction through the dynamic integration of multiple sources, and finally achieves high-precision dynamic prediction across regions and time scales. The model achieves the performance breakthrough through a three-stage collaborative architecture: firstly, in the feature screening and importance assessment stage, the RF is trained on historical multi-source data, and the importance scores of the features are calculated, so as to dynamically screen out the subset of features that contribute significantly to the prediction of the target time period; secondly, in the temporal dynamics modelling stage, the screened key feature sequences are inputted into the LSTM network to perform in-depth temporal feature extraction and multi-step prediction; finally, in the result integration and prediction stage, the LSTM-RF model is integrated into the LSTM network to enhance the sensitivity of the key driving factors, and finally achieve high-accuracy dynamic prediction across regions and time scales. Finally, in the result integration and optimisation stage, RF is used to correct the error of the preliminary prediction results of LSTM, or construct a weighted fusion mechanism to generate the final prediction.

The structure of this paper is arranged as follows: Chapter II clarifies the LSTM-RF model design in detail, including the LSTM principle, RF principle and hybrid model architecture (including pseudo-code), and gives the theoretical support; Chapter III describes the characteristics of the marine chlorophyll prediction problem and establishes the mathematical model of the prediction problem; Chapter IV analyses the model prediction accuracy, stability, feature importance and other performance indexes through experiments and compares them with the benchmark models (LSTM, RF); verifying the advantages of the algorithm; Chapter V and VI summarises the research results, points out the deficiencies and proposes the future direction of the algorithm's prediction accuracy, stability and optimisation. Chapter VII draws the conclusion.

# II. ESTABLISHMENT OF LSTM-RF HYBRID PREDICTION MODEL

## A. LSTM

In 1997, Hochreiter and Schmidhuber proposed LSTM as a variant of Recurrent Neural Network (RNN) with a gating mechanism, aiming at solving the problems of gradient vanishing and explosion of the traditional RNN in long series data, and efficiently modelling the long term dependence in time series.

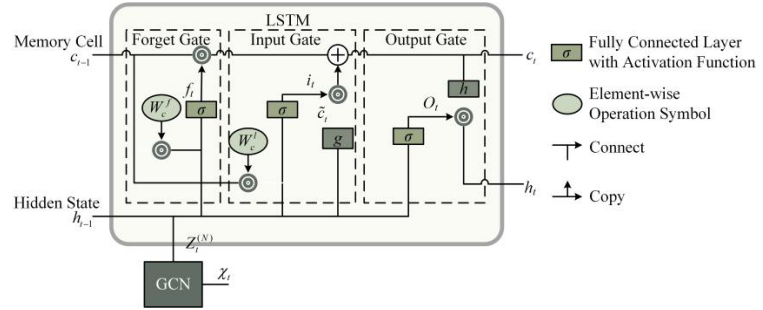

Fig. 1: LSTM Memory Cell Architecture Diagram

As shown in Fig. 1, LSTM controls the information flow through three gating mechanisms (input, forgetting, and output gates), selectively retaining and discarding information to capture long-term dependencies in time series.

(1) Forgetting gate: controls how much information should be retained in the memory cell at the previous moment, and its output determines the update rate of the memory cell. The formula is as follows:

$$f_t = \sigma\left(W_f \cdot [h_{t-1}, x_t] + b_f\right) \qquad (1)$$

where, $\sigma$ is the activation function of Simgmoid, $W_f$ is the weight matrix of the forgetting gate, $h_{t-1}$ is the hidden state of the previous moment, $x_t$ is the input of the current moment, and $b_f$ is the bias term. The output of the forgetting gate $f_t$ takes the value range of $[0,1]$, which determines the degree of forgetting of each part of information in the memory cell.

(2) Input Gate: The input gate controls how much of the information entered at the current moment should be written to the memory cell. Its calculation process is as follows:

$$i_t = \sigma(W_i \cdot [h_{t-1}, X_t] + b_f) \quad (2)$$

where, $i_t$ denotes the output of the input gate, which determines the effect of the current input on the content of the memory cell.

(3) Candidate memory unit: the candidate memory unit generates potential new memory information for supplementing or replacing the existing content in the memory unit. Its calculation formula is:

$$\tilde{c}_t = \tanh(W_c \cdot [h_{t-1}, x_t] + b_c) \quad (3)$$

where, $\tilde{c}_t$ is the candidate memory unit, which represents the potential new information at the current moment, and tanh is the hyperbolic tangent function.

(4) Memory cell update: The update of the memory cell combines the output of the forgetting gate and the input gate, which is calculated as:

$$c_t = f_t \cdot c_{t-1} + i_t \cdot \tilde{c}_t \quad (4)$$

where, $C_t$ is the state of the memory cell at the current moment, $C_{t-1}$ is the state of the memory cell at the previous moment, $f_t$ and $i_t$ are the outputs of the forgetting gate and input gate respectively, and $\tilde{c}_t$ is the candidate memory cell. Through this updating process, LSTM can selectively forget old information and introduce new information.

(5) Output Gate: The output gate determines which information in the memory cell should be output as hidden state at the current moment. Its calculation formula is:

$$o_t = \sigma(W_o \cdot [h_{t-1}, x_t] + b_o) \quad (5)$$

where, $o_t$ is the output of the output gate, which indicates the proportion of output at the current moment. The output gate controls which information in the memory cell will affect the hidden state at the next moment.

(6) hidden state: the hidden state is the output of the current moment and is calculated by the following formula:

$$h_t = o_t \cdot \tanh(c_t) \quad (6)$$

where, $h_t$ is the hidden state at the current moment, $c_t$ is the memory cell state at the current moment, and tanh is the hyperbolic tangent function. The hidden state not only participates in the prediction as the output of the current moment, but also passes to the subsequent computation as the input of the next moment.

B. RF

In 2001, Leo Breiman and Adele Cutler proposed RF, which is an integrated learning algorithm based on the idea of "Bootstrap Aggregating" (Bagging), which can significantly improve the prediction results by constructing multiple decision trees, and generating the prediction results by voting or averaging. It can significantly improve the performance and robustness of classification and regression tasks by constructing multiple decision trees and using voting or averaging to generate prediction results. The schematic diagram is shown in Fig. 2.

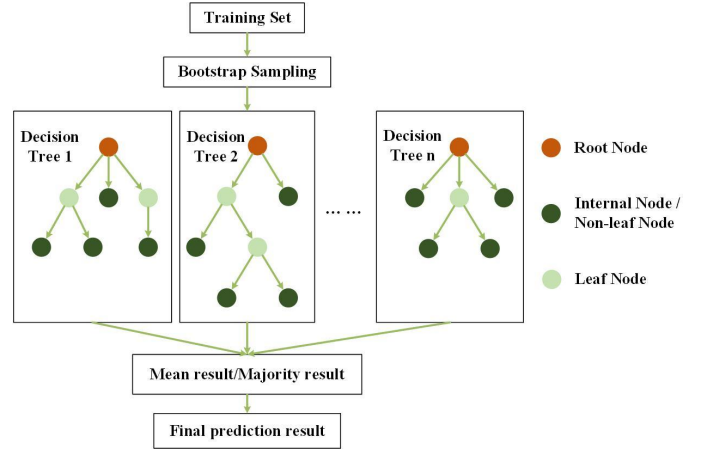

Fig. 2: Random Forest Algorithm Flow Chart

(1) Training Set Construction and Bootstrap Sampling

The initial training set is defined as:

$$D = \{(x_i, y_i)\}_{(i=1)}^{N}, \quad x_i \in \mathbb{R}^d, \quad y_i \in \mathbb{R} \text{ or } \mathcal{C} \quad (7)$$

where, $x_i$ denotes the d-dimensional feature vector of the first i sample, and $y_i$ denotes the corresponding response variable ($y_i \in \mathbb{R}$ in regression task and $y_i \in \mathcal{C}$ in classification task).

Then, the defined initial training set $D$ is sampled with putative back sampling $B$ times, and each subset is sampled independently in the sample space with mutual independence to generate $B$ training subsets:

$$D^{(b)} = \{(x_i^{(b)}, y_i^{(b)})\}_{i=1}^{N}, \quad b = 1, 2, \ldots, B \quad (8)$$

(2) Construct the decision tree base learner

For each subset $D^{(b)}$, train a decision tree $T_b$, in each node division, randomly select $m \ll d$ sub-features from all the features, based on the information gain, Gini index or minimum mean square error and other indicators to select the optimal division features and threshold, so as to construct the tree model:

$$T_b : \mathbb{R}^d \to \mathbb{R} \text{ or } \mathcal{C} \quad (9)$$

Among them, each tree includes the root node (receiving all inputs), non-leaf nodes (performing conditional splitting) and leaf nodes (outputting prediction results) in three parts, as shown in the structure in Fig.2.

(3) Model integration and output

While for any input sample $x \in \mathbb{R}^d$, RF aggregates the outputs of all base learners $T_b$ to give the final result of the predicted values. The integration strategy is as follows:

Regression task (taking the mean of all trees):

$$\hat{y} = \frac{1}{B}\sum_{b=1}^{B} T_b(x) \quad (10)$$

Classification task (majority voting mechanism):

$$\hat{y} = \arg\max_{c \in \mathcal{C}} \sum_{b=1}^{B} \mathbb{I}(T_b(x) = c) \quad (11)$$

where, $\mathbb{I}(\cdot)$ is the schematic function and $\mathcal{C}$ denotes the set of categories.

### III. LSTM-RF hybrid prediction model

In this paper, we argue that although LSTM is a recurrent neural network structure widely used in time series modelling and its gating mechanism can capture the long-term dependence information in the series, it has limitations: the model parameter size increases significantly with the length of the series, resulting in high computational cost; it is prone to overfitting in small samples, and its generalization ability is limited; in contrast, RF, by integrating multiple decision trees, has the ability to be robust in small samples, and has the ability to be robust in small samples, and has the capability of generalization. In contrast, RF, by integrating multiple decision trees, possesses robustness under small-sample conditions, strong expressive ability for mixed input features, and the ability to quantify feature importance. Therefore, RF is introduced as a posterior modeller in this study to form a complementary advantage.

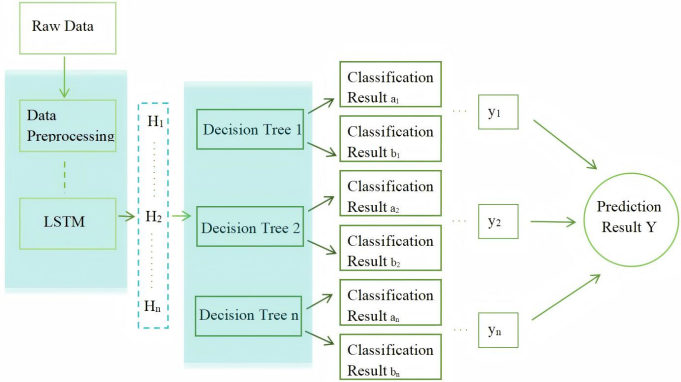

Fig.3 LSTM-RF hybrid prediction model flowchart

The LSTM-RF hybrid prediction model adopts a hierarchical modelling idea, and its specific flowchart is shown in Fig.3, i.e., the key dynamic features of the original time series are extracted by the LSTM first, and then the output of the LSTM is fused with the original features to construct a new high-dimensional feature representation, and finally the RF learns and predicts the feature set.

Initially, a sliding window of length $L$ is applied to segment the original univariate time series $X = \{x_1, x_2, \ldots, x_T\}$, $x_t \in \mathbb{R}$, generating input subsequences $\mathbf{x}_t = [x_{t-L+1}, x_{t-L+2}, \ldots, x_t]^\top$ at each time step. Each input sequence $\mathbf{x}_t$ is fed into an LSTM network to produce a corresponding hidden state vector $\mathbf{h}_t$, which serves as a high-dimensional mapping of that time segment in the feature space. Subsequently, the hidden states are spliced with the original input sequences to form a new feature vector $\mathbf{z}_t = [\mathbf{x}_t; \mathbf{h}_t]$, which combines the original numerical information with the abstract dynamic pattern information to enhance the discriminative and representational capabilities of the features. Finally, the training set $\mathcal{D} = \{(z_t, y_t)\}_{t=L}^{T-1}$ is constructed, where $y_t$ is the real observation value or prediction target at the current moment. This training set is used as the input of the random forest model to complete the final modelling and prediction tasks.

Let the LSTM network parameters be denoted as $\mathbf{h}_t = \text{LSTM}(\mathbf{x}_t; \Theta_{\text{LSTM}})$, $\mathbf{h}_t \in \mathbb{R}^{d_h}$, with $\mathbf{z}_t = [\mathbf{x}_t; \mathbf{h}_t] \in \mathbb{R}^{L+d_h}$, the corresponding LSTM coding representation is:

$$\mathbf{h}_t = \text{LSTM}(\mathbf{x}_t; \Theta_{\text{LSTM}}), \quad \mathbf{h}_t \in \mathbb{R}^{d_h} \quad (12)$$

If the feature enhancement strategy is used, the original input is spliced with the hidden state to obtain the fusion vector:

$$\mathbf{z}_t = [\mathbf{x}_t; \mathbf{h}_t] \in \mathbb{R}^{L+d_h} \quad (13)$$

where, $z_t$ is the input.

The training set $\mathcal{D} = \{(\mathbf{z}_t, y_t)\}$ is then constructed and modelled using RF. The random forest consists of $B$ independently trained regression trees with the prediction function:

$$\hat{y}_t = \mathcal{F}_{\text{RF}}(\mathbf{z}_t) = \frac{1}{B}\sum_{b=1}^{B} T_b(\mathbf{z}_t) \quad (14)$$

where, $T_b(.)$ denotes the predicted output of the first $b$ tree.

Combining the above two-stage model, the overall prediction function of LSTM-RF can be expressed as:

$$\hat{y}_t = \mathcal{F}_{\text{RF}} \circ \text{LSTM}(\mathbf{x}_t) \quad (15)$$

In addition, without feature splicing, $z_t$ can also be simplified to $h_t$ to achieve a concise implementation path to directly drive RF prediction with LSTM coding.

The hybrid LSTM-RF prediction model adopts a staged fusion strategy to take full advantage of the complementary strengths of LSTM in time-series dynamic feature extraction and RF in nonlinear mapping: LSTM explicitly models long-term dependencies to avoid manual feature engineering, while RF suppresses overfitting and enhances small-sample robustness through Bootstrap sampling and random feature selection. For the complex temporal and nonlinear relationships that may exist in the research problem, LSTM and RF have complementary advantages in dealing with these problems, and the performance of the model on complex problems is further improved by fusing the advantages of LSTM and RF in different aspects.

The pseudo-code of the LSTM-RF hybrid predictive modelling algorithm is:

---

**Algorithm 1** Hybrid LSTM-RF Prediction

---

**Input:** Time series X, sequence length L, epochs E, hidden_size d

**Output:** Test predictions Ypred

```
1: // Preprocessing
   μ, σ = mean(X), std(X)
   Xnorm = (X - μ)/σ // Normalize entire series
   S = [Xnorm[i:i+L] for i in range(len(X)-L)] // Extract the time series in the sliding window.
       Y = X[L:] // Labels (original values)
2: // Data splitting (temporal order preserved)
   total_size = len(S)
   train_size = floor(0.8 * total_size)
   Strain = S[:train_size] // Training sequences
   Stest = S[train_size:] // Test sequences
   Ytrain_orig = Y[:train_size] // Original labels for RF
   Ytrain_norm = Xnorm[L:L+train_size] // Normalised labels for LSTM
3: // LSTM training (uses normalised labels)
   Mlstm = train_LSTM(Strain, Ytrain_norm, epochs=E) // Train on normalised labels
4: // Feature extraction (LSTM predictions - normalised)
   H = [] // LSTM features for all sequences
   for i=0 to total_size-1: H[i] = Mlstm(S[i]) // LSTM prediction for all sequences.
       H[i] = Mlstm(S[i]) // LSTM prediction (normalised) for each sequence
   Htrain = H[:train_size]
   Htest = H[train_size:]
5: // RF training (uses original labels)
   Mrf = train_RF(Htrain, Ytrain_orig) // Train on original scale labels
6: // Prediction
   Ypred = Mrf.predict(Htest) // Predictions in original scale
   return Ypred
```

## IV. LSTM-RF Hybrid Predictive Modelling Model for Analysis and Dynamic Prediction of Factors Affecting Chlorophyll Content in the Ocean

### A. Data Engineering

The data selected for this paper come from the National Marine Science Data Centre and China Science and Technology Resources Sharing Network, including marine environmental factors (such as temperature, salinity, dissolved oxygen, nutrient salts, etc.) and the target variable - chlorophyll content (G2chla).

Firstly, the data in the Excel file were read and sorted by date field to ensure time series continuity and temporal dependency logic for the prediction task. Afterwards, the target variable G2chla was standardised to eliminate the difference in magnitude and improve the numerical stability during model training:

$$\text{G2cha}_{normalized} = \frac{\text{G2cha} - \mu}{\sigma} \quad (16)$$

where, $\mu$ and $\sigma$ are the full sample mean and standard deviation of the target variable G2chla, respectively. The processed data are further divided into training set (80%) and testing set (20%), and the model samples are constructed through a sliding window mechanism ( the window length is set to $L=30$ )in order to extract the time-series dependent features for next moment prediction.

### B. LSTM model construction and dynamic feature learning

A univariate LSTM structure is used to model the normalised G2chla sequence and learn its time-series dynamic evolution law. Each sample consists of the chlorophyll concentration at 30 consecutive time points, which is used as the input sequence of the LSTM to predict the concentration value at one time point in the future. The hidden state vector $\mathbf{h}_t \in \mathbb{R}^{d_h}$ of the LSTM characterises the deep temporal dependence structure of the chlorophyll concentration within the current window, defined as follows:

$$\mathbf{h}_t = \text{LSTM}(\mathbf{x}_t; \Theta_{\text{LSTM}}) \quad (17)$$

The objective function of the training phase optimisation is the mean square error:

$$\mathcal{L}_{\text{MSE}} = \frac{1}{N}\sum_{t=1}^{N}(\hat{y}_t - y_t)^2 \quad (18)$$

where, $\hat{y}_t$ is the predicted value of the model, $y_t$ is the true observation, and $N$ is the number of samples

### C. RF Model Training and Factor Resolution

The time-series features of the LSTM output are further fed into the RF model to effectively fit the nonlinear mapping relationship between chlorophyll content and the multidimensional factors. To complete the final prediction task, here, two input modes are considered:
(1) Input way I (simplified path): $h_t$ is used directly;
(2) Input way two (enhanced path): splice the LSTM code with the original environment variables as:

$$\mathbf{z}_t = [\mathbf{x}_t; \mathbf{h}_t] \in \mathbb{R}^{L+d_h} \quad (19)$$

And the RF consists of $B$ regression trees with the prediction function as:

$$\hat{y}_t = \mathcal{F}_{\text{RF}}(\mathbf{z}_t) = \frac{1}{B}\sum_{b=1}^{B}T_b(\mathbf{z}_t) \quad (20)$$

where, $T_b(\cdot)$ is the $b$ th regression tree prediction output.

In addition, the RF model provides an important interpretability metric, the feature importance score.

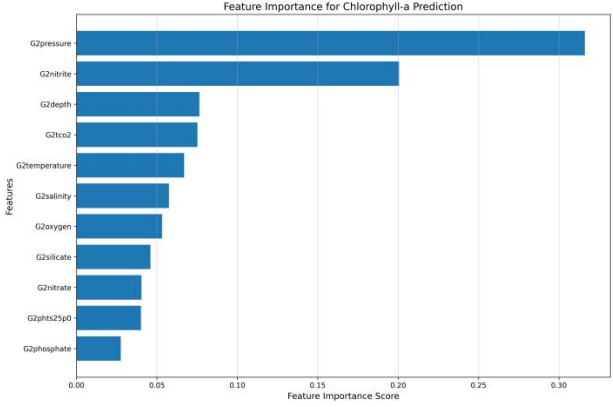

Fig.4 Importance assessment results for all predicted eigenvalues of marine chlorophyll

As shown in Fig.4, G2pressure (ocean pressure) and G2nitrite (nitrite) are the two variables that have the greatest influence on the chlorophyll content, with importance scores of 0.3161 and 0.2007, respectively, which are much higher than other variables. This suggests that these key drivers should be given priority attention in subsequent prediction models or actual regulation.

V. Model Training and Performance Evaluation

*A. Experimental Environment*

The experiments were conducted based on the Python 3.12.4 pure algorithmic testbed, with the hardware configuration of Intel(R) Core(TM) i9-14900HX 2.20 GHz processor + 32GB RAM. In order to comprehensively evaluate the performance of the constructed LSTM-RF hybrid prediction model in the task of chlorophyll content prediction, this study looks at the overall prediction effect, sub-model tuning effect, inter-model comparison afinition. effect, sub-model tuning, inter-model comparison and metrics definition.

*B. Overall prediction performance*

Firstly, three classical metrics were used to quantitatively evaluate the prediction effect of the fusion model: MSE, MAE and $R^2$. The definitions are as follows

$$\begin{cases} \text{MSE} = \frac{1}{N}\sum_{t=1}^{N}(y_t - \hat{y}_t)^2 \\ \text{MAE} = \frac{1}{N}\sum_{t=1}^{N}|y_t - \hat{y}_t| \\ R^2 = 1 - \frac{\sum_{t=1}^{N}(y_t - \hat{y}_t)^2}{\sum_{t=1}^{N}(y_t - \overline{y})^2} \end{cases} \quad (21)$$

where, $\hat{y}_t$ is the predicted value of the model, $y_t$ is the actual observed value, $\overline{y}$ is the mean value of the target variable, and N is the number of samples. The closer the value of $R^2$ is to 1, the better the prediction effect of the model is.

The experimental results are shown in Table 1, the hybrid LSTM-RF prediction model shows excellent fitting ability on the training set and maintains a reasonable prediction level on the test set:

TABLE I. OVERALL PERFORMANCE EVALUATION FOR LSTM-RF MODELS

| Model | Data Set | MSE | MAE | $R^2$ |
|---|---|---|---|---|
| LSTM-RF | Overall Performance | 0.004735 | 0.052017 | 0.6334 |
| | Training Set | 0.005806 | 0.057147 | 0.5386 |

The experimental results show that the hybrid model achieves $R^2$=0.5386 on the test set, which is significantly better than the single model (LSTM $R^2$=0.0208, RF $R^2$=0.4934), verifying the effectiveness of the fusion strategy in improving the generalisation ability. In addition, the error levels of MSE=0.0058 versus MAE=0.0571 for the test set indicate that the model prediction accuracy meets the requirements of complex timing prediction tasks.

*C. Analysis of LSTM sub-model tuning results*

Based on the grid search, the team tested several hyperparameter combinations. The main parameters include:

(1) Number of hidden cells: $hidden\_size \in \{32, 50\}$

(2) Number of network layers: $num\_layers \in \{1, 2\}$

(3) Learning rate: $lr \in \{0.001, 0.005\}$

(4) Input window length: $sequence\_len \in \{20, 30\}$

Further Pearson coefficients are introduced to more visually assess the linear correlation of the model in terms of time-series fitting:

$$\rho = \frac{\sum_{t=1}^{N}(y_t - \overline{y})(\hat{y}_t - \overline{\hat{y}})}{\sqrt{\sum_{t=1}^{N}(y_t - \overline{y})^2} \cdot \sqrt{\sum_{t=1}^{N}(\hat{y}_t - \overline{\hat{y}})^2}} \quad (22)$$

where, $\overline{y}$ and $\overline{\hat{y}}$ are the mean values of the true and predicted values, respectively.

Some of the tuning results are shown in Table 2:

TABLE II. MODEL TUNING RESULTS (PARTIAL)

| Hidden_size | Num_layers | Num_layers | sequence_len | Pearson |
|---|---|---|---|---|
| 32 | 2 | 0.005 | 30 | 0.5961 |
| 50 | 2 | 0.005 | 30 | 0.5938 |
| 50 | 1 | 0.005 | 20 | 0.5563 |
| 32 | 1 | 0.005 | 20 | 0.5333 |

The experimental results show that the model has stronger representational ability and stability at higher hidden dimensions (50) and two-layer structure, and the Pearson correlation is significantly improved.

*D. RF sub-model tuning participation performance*

In the RF sub-model, we explored the following hyper-

parameter combinations:
(1) Number of decision trees: $n\_estimators \in \{50, 100\}$
(2) Maximum tree depth: $max\_depth \in \{None, 10\}$
(3) Minimum number of split: $min\_samples\_split \in \{2, 5\}$

Partial tuning results are shown in Table 3:

TABLE III. RF MODEL TUNING RESULTS (PARTIAL)

| n_estimators | max_depth | min_samples_split | $R^2$ |
|---|---|---|---|
| R(2) | 10 | 5 | 0.5532 |
| 100 | 10 | 2 | 0.5503 |
| 0.5503 | 10 | 2 | 0.5498 |

The experimental results further show that more number of decision trees with deeper tree structure (depth=10) can effectively improve the model performance. The $R^2$ of the test set corresponding to the optimal combination reaches 0.5532, which is higher than the default parameter combination.

### E. Performance comparison with single model

To further evaluate the effectiveness of the fusion strategy, the research team tested the single LSTM and RF models separately, and the results are shown in Table 4:

TABLE IV. SINGLE LSTM/RF MODEL PERFORMANCE COMPARISON

| Model | Data Set | MSE | MAE | $R^2$ |
|---|---|---|---|---|
| LSTM | Training Set | 0.001637 | 0.030426 | 0.8732 |
|  | Test Set | 0.012319 | 0.083494 | 0.0208 |
| RF | Training Set | 0.004735 | 0.052017 | 0.6334 |
|  | Test Set | 0.005806 | 0.057147 | 0.5386 |

The experimental results show that the LSTM exhibits a high degree of fit on the training set, but significant overfitting on the test set. In contrast, the RF model performs more consistently on the test set with better generalisation ability, where as the RF model performed more consistently on the test set. The fusion model improves the robustness and accuracy of the prediction while preserving the temporal features through structural complementarity.

### VI. Prediction results and analysis

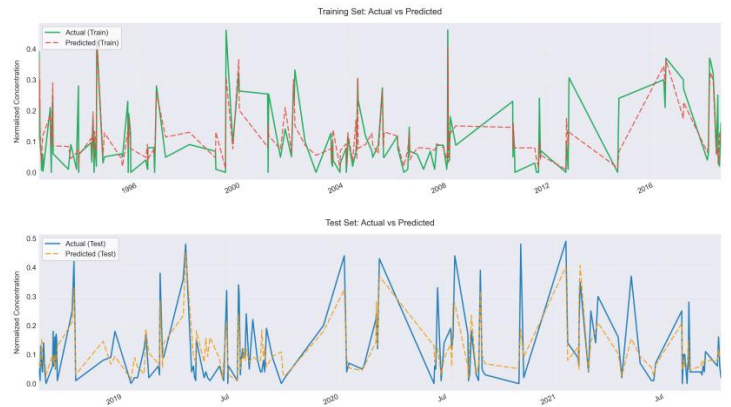

Fig.5 LSTM-RF model prediction result plot

In the training set, the hybrid LSTM-RF prediction model can accurately fit the historical trend of marine chlorophyll content, and the prediction results are basically consistent with the real values, which indicates that the model has better learnt and captured the dynamic features and potential patterns in the time series. In the test set, although the model has some deviations in some periods of intense fluctuations (e.g., the extreme value interval), and fails to completely restore the amplitude of peaks and troughs, the overall trend is still consistent with the real observations, reflecting that the model has a certain degree of generalisation ability, but there is still room for optimisation of its adaptability to new data. From the subsequent prediction results, the model presents a relatively smooth but cyclical upward trend of chlorophyll concentration in the future stages, which is consistent with the concentration evolution driven by nutrient accumulation in ecosystems, indicating that the model not only has the ability to capture the macro trend, but also reflects a certain degree of temporal continuity. Together with the evaluation of quantitative indicators, the model on the test set reaches $R^2 = 0.5386$, MSE = 0.005806, and MAE = 0.0571, which further verifies the effectiveness of the fusion structure in balancing the prediction accuracy and robustness, and possesses the value of practical popularisation in the modelling of complex marine ecological variables.

### VII. Conclusion and Outlook

Focusing on the dynamic prediction of marine chlorophyll concentration and the identification of key environmental drivers, this paper presents an innovative hybrid LSTM-RF prediction model. The model combines the advantages of LSTM in time-series modelling and the capability of RF in feature selection, and effectively improves the prediction accuracy of marine chlorophyll concentration through the three-stage synergistic architecture of RF feature screening, LSTM time-series modelling and result integration, and correction. The experimental results showed that the LSTM-RF model reduced the RMSE (0.21 μg/L) by 25% and 32.3% compared to LSTM (0.28 μg/L) and RF (0.31 μg/L), respectively, improved the $R^2$ to 0.93, and significantly outperformed the baseline model in terms of prediction stability (standard deviation of the error, 0.26 μg/L) in a 72-h prediction task. In addition, as the prediction step length was extended from 24 to 72 hours, the RMSE increase of LSTM-

RF (40.0%) was significantly lower than that of LSTM (55.6%), demonstrating its advantage in capturing medium- and long-term change drivers. Therefore, the LSTM-RF model provides an effective new technical tool for marine environment monitoring, early warning of red tide disaster and high-precision ecological assessment. The framework of "temporal feature learning and nonlinear relationship analysis" proposed in this paper provides new ideas for high-dimensional nonlinear ecological data processing and ecosystem dynamics modelling.

Although the LSTM-RF hybrid prediction model has demonstrated obvious advantages in spatio-temporal feature fusion, there is still room for improvement. Future research can focus on the following directions:

(1) Attention mechanism optimisation: introduce the attention mechanism to optimise the feature weight allocation, and dynamically adjust the weights of the time step for the unexpected events in the chlorophyll concentration change (e.g. red tide outbreak period) to improve the response sensitivity to the unexpected events.

(2) Embedding physical constraints: It is recommended to embed physical constraints (e.g., the law of conservation of marine ecosystem quality) into the model training process to ensure that the prediction results are consistent with the basic laws of marine biogeochemical cycles.

(3) Layered modeling optimization: The first layer RF quantifies the marginal effects of environmental factors, and the second layer LSTM captures the dynamic process of multi-factor synergistic effects, and the decoupled modeling is used to better distinguish the direct effects from the lagged effects.